\documentclass[10pt,twocolumn,letterpaper]{article}

\usepackage[pagenumbers]{cvpr}      

\usepackage[dvipsnames]{xcolor}

\usepackage{amsmath,amsfonts,bm}

\def\eqref#1{equation~\ref{#1}}

\def\1{\bm{1}}

\DeclareMathAlphabet{\mathsfit}{\encodingdefault}{\sfdefault}{m}{sl}
\SetMathAlphabet{\mathsfit}{bold}{\encodingdefault}{\sfdefault}{bx}{n}

\usepackage{url}
\usepackage{multirow}

\usepackage{xparse}
\usepackage{pifont}
\usepackage{bm}
\usepackage{xspace} 

\newcommand{\cmark}{\ding{51}}%
\newcommand{\xmark}{\ding{55}}%

\definecolor{cvprblue}{rgb}{0.21,0.49,0.74}
\usepackage[pagebackref,breaklinks,colorlinks,citecolor=cvprblue]{hyperref}

\def\paperID{} 
\def\confName{CVPR}
\def\confYear{2024}

\title{Fast Training of Diffusion Transformer with Extreme Masking \\ for 3D Point Clouds Generation}

\author{
Shentong Mo$^1$~ Enze Xie$^{2}$\thanks{Corresponding author.}~~ Yue Wu$^2$~ Junsong Chen$^{2}$~
  Matthias Nießner$^3$~ Zhenguo Li$^2$
  \\
  $^1$MBZUAI, $^2$Huawei Noah's Ark Lab, $^3$TUM \\
  \url{https://DiT-3D.github.io/FastDiT-3D} \\
}

\begin{document}
\maketitle

\begin{abstract}

Diffusion Transformers have recently shown remarkable effectiveness in generating high-quality 3D point clouds. 
However, training voxel-based diffusion models for high-resolution 3D voxels remains prohibitively expensive due to the cubic complexity of attention operators, which arises from the additional dimension of voxels.
Motivated by the inherent redundancy of 3D compared to 2D, we propose FastDiT-3D, a novel masked diffusion transformer tailored for efficient 3D point cloud generation, which greatly reduces training costs.
Specifically, we draw inspiration from masked autoencoders to dynamically operate the denoising process on masked voxelized point clouds. We also propose a novel voxel-aware masking strategy to adaptively aggregate background/foreground information from voxelized point clouds. Our method achieves state-of-the-art performance with an extreme masking ratio of nearly 99\%.
Moreover, to improve multi-category 3D generation, we introduce Mixture-of-Expert (MoE) in 3D diffusion model. Each category can learn a distinct diffusion path with different experts, relieving gradient conflict.
Experimental results on the ShapeNet dataset demonstrate that our method achieves state-of-the-art high-fidelity and diverse 3D point cloud generation performance. Our FastDiT-3D improves 1-Nearest Neighbor Accuracy and Coverage metrics when generating 128-resolution voxel point clouds, using only 6.5\% of the original training cost.

\end{abstract}

\section{Introduction}

\begin{figure}[t]
\centering
\includegraphics[width=0.95\linewidth]{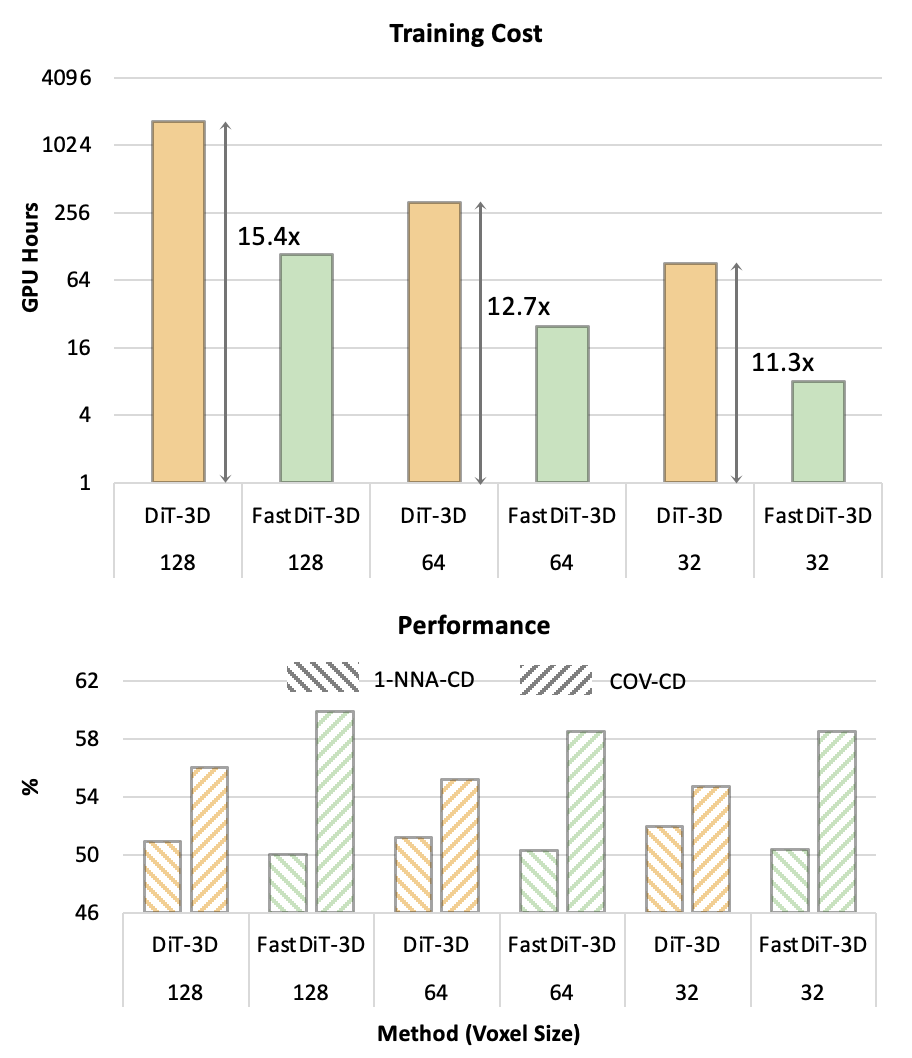}
\caption{Comparison of the proposed FastDiT-3D with DiT-3D in terms of different voxel sizes on training costs (lower is better) and COV-CD performance (higher is better).
Our method achieves faster training while exhibiting superior performance.
}
\label{fig: title_img}
\end{figure}

Recent breakthroughs in Diffusion Transformers have made remarkable strides in advancing the generation of high-quality 3D point clouds.
Notably, the current state-of-the-art (SOTA), DiT-3D~\cite{mo2023dit3d}, leveraged a diffusion transformer architecture for denoising voxelized point clouds, significantly outperformed previous UNet-based methods such as LION~\cite{zeng2022lion} by improving 1-Nearest Neighbor Accuracy (1-NNA) at 8.49\% and Coverage (COV) at 6.51\% in terms of Chamfer Distance (CD).
They also achieved superior performance compared to the previous best UNet-based mesh generation model MeshDiffusion~\cite{liu2023meshdiffusion}. Based on their excellent experimental results, adopting transformer architecture is expected to be the mainstream approach for 3D shape generation tasks.
Despite their efficacy, the voxel-based diffusion transformer's training overhead significantly increases primarily due to the additional dimension when transferring from 2D to 3D. This results in cubic complexity associated with attention mechanisms within the volumetric space. 
For instance, training voxels of 128 × 128 × 128 takes 1,668 A100 GPU hours. Such a large amount of computational resources is the bottleneck to further increasing the input voxel size or scaling up these model architectures. The training efficiency of diffusion transformers in 3D shape generation is still an unsolved problem.

In image generation and visual recognition, masked training~\cite{he2022masked,chang2023muse,chang2022maskgit,zheng2023fast} is widely adopted to improve training efficiency, which significantly reduces training time and memory but does not comprise the performance. Considering the high redundancy of 3D voxels, only a partial of the volumetric space is occupied. It is possible to generate high-fidelity 3D shape training on a subset of voxels.

In this work, we introduce FastDiT-3D, a novel diffusion transformer architecture explicitly designed to generate 3D point clouds efficiently. 
Inspired by masked autoencoders~\cite{he2022masked}, we propose a dynamic denoising operation on selectively masked voxelized point clouds. We further propose a novel foreground-background aware masking strategy, which adaptly aggregates information by differentiating between the information-rich foreground and information-poor background within the point clouds. This innovative approach achieves an outstanding masking ratio, with almost 98\% of input voxels masked, superior to the 50\% observed in 2D~\cite{zheng2023fast}, leading to a remarkable 13X acceleration in training speed. Moreover, to address the heightened computational demands posed by the increased token length in 3D contexts, we integrate 3D window attention mechanisms within the decoder's Transformer blocks. 
Our training regimen employs a dual-objective strategy, applying a denoising objective to unmasked patches while masked patches undergo a distinct point cloud generation objective. Our approach not only accelerates the training process but also achieves SOTA performance.

To enhance the capability of point cloud generation across diverse categories, we incorporate Mixture of Expert~(MoE) layers within the Transformer blocks.
In this way, we transform a dense 3D diffusion model into a sparse one. 
Each category can learn a distinct diffusion path, and each diffusion path is composed of different experts across different layers. 
This design greatly alleviates the challenge of difficult gradient optimization caused by multi-category joint training.

Our comprehensive evaluation on the ShapeNet dataset conclusively attests to FastDiT-3D's state-of-the-art performance in generating high-fidelity and diverse 3D point clouds across categories, evidenced by improved 1-NNA and COV metrics for 128-resolution voxel point clouds. 
Remarkably, our model achieves these results at a mere \textbf{6.5\%} of the original training cost. 
Qualitative visualizations further corroborate FastDiT-3D's proficiency in rendering detailed 3D shapes. 
A series of ablation studies underscore the critical roles played by the foreground-background aware masking, the encoder-decoder architecture, and the dual training objectives in the adept learning of our FastDiT-3D. 
Lastly, incorporating MoE distinctly showcases the model's effectiveness in accommodating multiple categories through a unified global model.

Our main contributions can be summarized as follows:
\begin{itemize}
    \item We present a fast diffusion transformer based on encoder-decoder architecture for point cloud shape generation, called FastDiT-3D, that can efficiently perform denoising operations on masked voxelized point clouds with an extreme masking ratio, which masks 99\% of the background and 95\% of the foreground.
    \item We propose a novel foreground-background aware masking mechanism to select unmasked patches for efficient encoding and Mixture of Expert (MoE) Feed-forward Network in encoder blocks for multi-category adaptation. 
    \item Comprehensive experimental results on the ShapeNet dataset demonstrate the state-of-the-art performance against the original DiT-3D while largely reducing the training costs.
\end{itemize}

\section{Related Work}

\noindent\textbf{3D Shape Generation.}
The domain of 3D shape generation primarily revolves around creating high-quality point clouds through the utilization of generative models.
These methods encompass various techniques, including variational autoencoders~\cite{Yang2018foldingnet,gadelha2018multiresolution,Kim2021SetVAE}, generative adversarial networks~\cite{valsesia2019learning, achlioptas2018learning,Shu2019pointcloud}, normalized flows~\cite{yang2019pointflow,Kim2020SoftFlowPF,Klokov2020dpfnet}, and Diffusion Transformers~\cite{mo2023dit3d}.

For example, Valsesia et al.~\cite{valsesia2019learning} proposed a generative adversarial network leveraging graph convolution. 
Klokov et al.~\cite{Klokov2020dpfnet} introduced a latent variable model that employed normalizing flows to generate 3D point clouds.
GET3D~\cite{gao2022get3d} used two latent codes to generate 3D signed distance functions (SDF) and textures, enabling the direct creation of textured 3D meshes.

Most recently, DiT-3D~\cite{mo2023dit3d} pioneered the integration of denoising diffusion probabilistic models in the realm of 3D point cloud generation. Its efficacy in producing high-quality 3D point clouds has set a new benchmark in this domain, showcasing state-of-the-art performance. 
However, training voxel-based diffusion models for high-resolution 3D voxels ($128 \times 128 \times 128 \times 3$) remains prohibitively expensive due to the cubic complexity of attention operators, which arises from the additional dimension of voxels. Our focus is to explore methods for expediting the training process while upholding the generation quality. This exploration is critical to mitigate the computational constraints without compromising the fidelity of the generated outputs.

\begin{figure*}[t]
\centering
\includegraphics[width=0.89\linewidth]{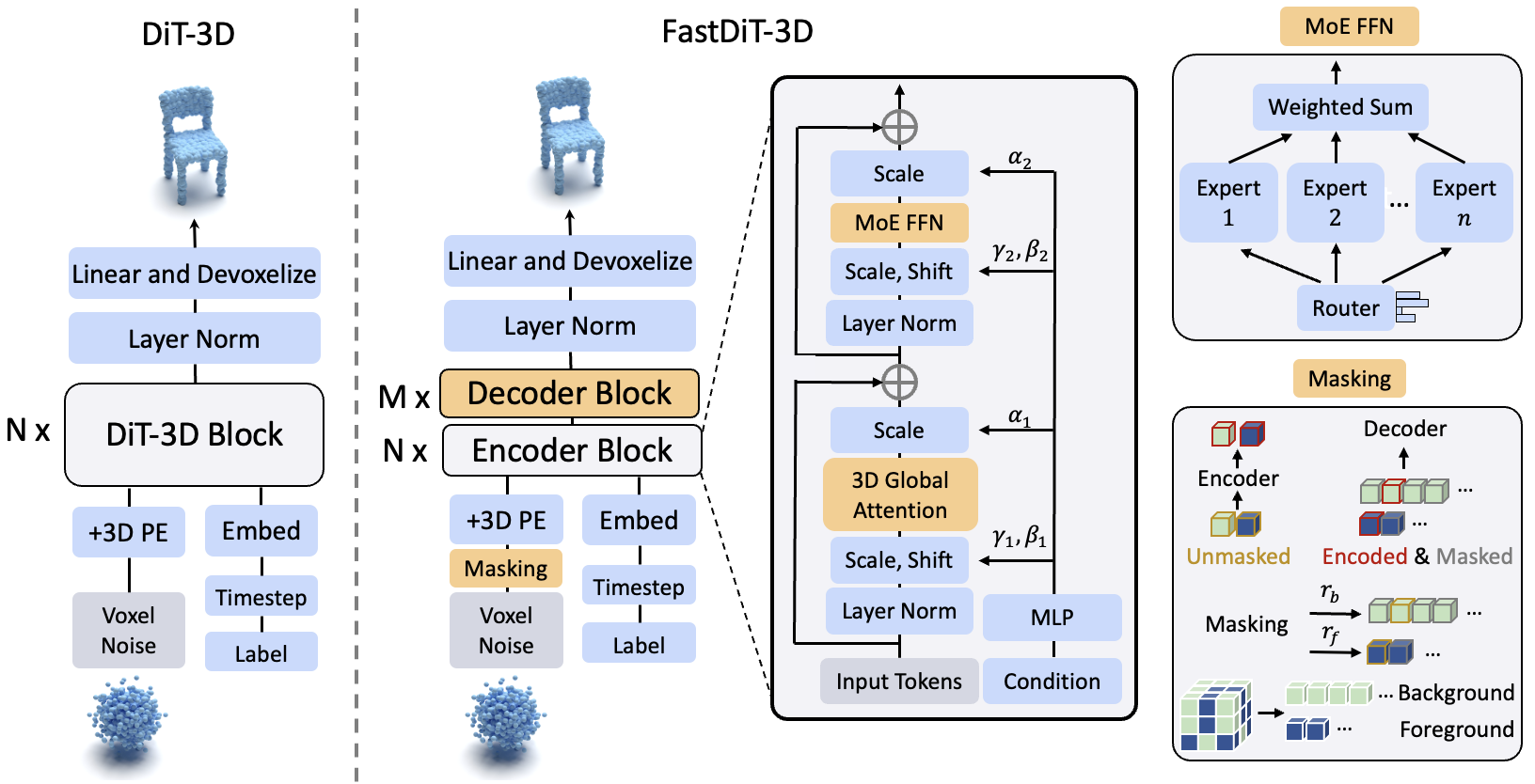}
\vspace{-0.5em}
\caption{Illustration of the proposed Fast training of Diffusion Transformers~(FastDiT-3D) for 3D shape generation.
The encoder blocks with 3D global attention and Mixture-of-Experts (MoE) FFN take masked voxelized point clouds as input. 
Then, multiple decoder transformer blocks based on 3D window attention extract point-voxel representations from all input tokens.
Finally, the unpatchified voxel tensor output from a linear layer is devoxelized to predict the noise in the point cloud space.
}
\label{fig: main_img}
\vspace{-1.0em}
\end{figure*}

\noindent{\textbf{Diffusion Transformers in 3D Point Clouds Generation.}}
Recent research, as documented in works such as~\cite{Peebles2022DiT, bao2022all, bao2023transformer, xie2023difffit}, has highlighted the impressive performance of Diffusion Transformers. Diffusion Transformers have exhibited remarkable proficiency in generating high-fidelity images and even 3D point clouds, as outlined in~\cite{mo2023dit3d}.
In the area of image generation, the Diffusion Transformer (DiT)~\cite{Peebles2022DiT} presented a plain diffusion Transformer architecture aimed at learning the denoising diffusion process on latent patches. 
The U-ViT model~\cite{bao2022all} employed a Vision Transformer (ViT)~\cite{Dosovitskiy2021vit}-based architecture with extensive skip connections.

In 3D point cloud generation, DiT-3D~\cite{mo2023dit3d} presented a novel plain diffusion transformer tailored for 3D shape generation, specifically designed to perform denoising operations on voxelized point clouds effectively. This method achieved state-of-the-art performance and surpassed previous GAN-based or normalized flows-based methods by a large margin, demonstrating the effectiveness of diffusion transformer architecture in the 3D point cloud generation. However, it is worth noting that the training process is computationally expensive, prompting the exploration of methods to expedite and optimize the training phase.

\noindent{\textbf{Mask Diffusion Transformers.}}
Transformers have emerged as predominant architectures in both natural language processing~\cite{vaswani2017attention,devlin2018bert} and computer vision~\cite{dosovitskiy2020image,Peebles2022DiT}.
The concept of masked training has found widespread application in generative modeling~\cite{radford2018improving,chang2022maskgit,chang2023muse} and representation learning~\cite{devlin2018bert,he2022masked,li2022scaling}.
Within computer vision, a series of methodologies have adopted masked language modeling. MaskGiT~\cite{chang2022maskgit} and MUSE~\cite{chang2023muse} utilized the masked generative transformer for predicting randomly masked image tokens, enhancing image generation capabilities.
MAE~\cite{he2022masked} further shows masked autoencoders are scaleable self-supervised learners.
MDT~\cite{gao2023masked} introduced a mask latent modeling scheme and achieved $3 \times$ faster learning speed than DiT~\cite{Peebles2022DiT}.
MaskDiT~\cite{zheng2023fast} proposed an efficient approach to train large diffusion models with masked transformers by randomly masking out a high proportion of patches in diffused input images and achieves 31\% of the training time of DiT~\cite{Peebles2022DiT}. 
Our work is the first to exploit masked training in the 3D point cloud generation domain. 
Even for a voxel size of $32\times32\times 32$, our method achieves $10 \times$ faster training than the SOTA method DiT-3D~\cite{mo2023dit3d} while exhibiting superior performance.

\section{Method}

Given a set of 3D point clouds, we aim to learn a plain diffusion transformer for synthesizing new high-fidelity point clouds.
We propose a novel fast diffusion transformer that operates the denoising process of DDPM on masked voxelized point clouds, namely FastDiT-3D, which consists of two main modules: masked design DiT for 3D point cloud generation in Section~\ref{sec:mask} and Mixture-of-Experts encoder for multi-category generation in Section~\ref{sec:moe}.

\subsection{Preliminaries}
In this section, we first describe the problem setup and notations and then revisit DDPMs for 3D shape generation and diffusion transformers on 3D point clouds.

\noindent\textbf{Revisit DDPMs on 3D Shape Generation.}
In the realm of 3D shape generation, prior research, as exemplified by Zhou~\cite{zhou2021pvd,mo2023dit3d}, has leveraged DDPMs that involve a forward noising process and a reverse denoising process. 
In the forward pass, Gaussian noise is iteratively added to a real sample $\mathbf{x}_0$.
By utilizing the reparameterization trick, $\mathbf{x}_t$ can be expressed as $\mathbf{x}_t = \sqrt{\bar{\alpha}_t}\mathbf{x}_0 + \sqrt{1-\bar{\alpha}_t}\boldsymbol{\epsilon}$. $\boldsymbol{\epsilon}\sim \mathcal{N}(\mathbf{0},\mathbf{I})$, $\alpha_t = 1-\beta_t$, and $\bar{\alpha}_t = \prod_{i=1}^t \alpha_i$, indicating the noise magnitude. If the timestep $t$ is large, $\mathbf{x}_T$ would be a Gaussian noise.
For the reverse process, diffusion models are trained to optimize a denoising network parameterized by $\boldsymbol{\theta}$ to map a Gaussian noise into a sample gradually. 
The training objective can be formulated as a loss between the predicted noise generated by the model $\boldsymbol{\epsilon}_{\boldsymbol{\theta}}(\mathbf{x}_{t},t)$ and the ground truth Gaussian noise $\boldsymbol{\epsilon}$, denoted as $\mathcal{L}_{\text{simple}} = ||\boldsymbol{\epsilon}-\boldsymbol{\epsilon}_{\boldsymbol{\theta}}(\mathbf{x}_{t},t)||^2$. 

We train the diffusion model conditioned with class label, $p_{\boldsymbol{\theta}}(\mathbf{x}_{t-1}|\mathbf{x}_t, c)$. During inference, new point clouds can be generated by sampling a Gaussian noise $\mathbf{x}_{T}\sim \mathcal{N}(\mathbf{0},\mathbf{I})$, then gradually denoise to obtain a sample $\mathbf{x}_{0}$.

\noindent\textbf{Revisit DiT-3D on Point Clouds Generation.}
To address the generation challenge on inherently unordered point clouds, DiT-3D~\cite{mo2023dit3d} proposed to voxelize the point clouds into dense representation in the diffusion transformers to extract point-voxel features.
For each point cloud $\mathbf{p}_i\in\mathbb{R}^{N\times 3}$ with $N$ points for $x,y,z$ coordinates, DiT-3D first voxelized it as input $\mathbf{v}_i\in\mathbb{R}^{V\times V\times V\times 3}$, where $V$ denotes the voxel size.
Then, they applied the patchification operator with a patch size $p \times p \times p$ to generate a sequence of patch tokens $\mathbf{s}\in\mathbb{R}^{L\times3}$, where $L=(V/p)^{3}$ is the total number of patchified tokens.
Finally, several transformer blocks based on window attention were adopted to propagate point-voxel features.
To achieve the denoising process in the point cloud space, the unpatchified voxel tensor is devoxelized into the output noise $\boldsymbol{\epsilon}_{\boldsymbol{\theta}}(\mathbf{x}_{t},t) \in\mathbb{R}^{N\times 3}$.

Although DiT-3D~\cite{mo2023dit3d} achieved promising results in generating high-fidelity 3D point clouds, they take the whole number $L$ of patchified tokens as input to the encoder for feature propagation.
The training process is computationally expensive, prompting the exploration of methods to expedite and optimize the training phase.
Furthermore, the computational cost of 3D Transformers can be significantly high on the increased token length. Regarding high dimensions in 3D voxel space, such as $128\times 128\times 128$, the training cost will be 1,668 A100 GPU hours.
To address this challenge, we propose a novel fast plain diffusion transformer for 3D shape generation that can efficiently achieve the denoising processes on masked voxelized point clouds, as shown in Figure~\ref{fig: main_img}.

\subsection{DiT-3D for Masked Voxelized Point Clouds}\label{sec:mask}

\noindent\textbf{Motivation.}
In order to achieve an efficient denoising process using a plain diffusion transformer during training, we propose several masked 3D design components in Figure~\ref{fig: main_img} based on the SOTA architecture of DiT-3D~\cite{mo2023dit3d} for 3D point cloud generation. 
Specifically, we introduce a novel foreground-background-aware masking mechanism designed to mask voxelized point clouds as input. Such a novel strategy makes the masking ratio extremely high at nearly 99\%, effectively leveraging the high inherent redundancy present in 3D data. We also replace 3D window attention with 3D global self-attention in the encoder blocks to propagate point-voxel representations from all unmasked tokens and add multiple decoder blocks with 3D window attention to take all patches tokens to predict the noise in the point cloud space.
Finally, we apply a denoising objective on unmasked patches and a masked point cloud objective on masked patches for training our fast diffusion transformer on 3D point cloud generation.

\begin{table}[t]
	\renewcommand\tabcolsep{10.0pt}
    \renewcommand{\arraystretch}{1.1}
	\centering
	\scalebox{0.85}{
		\begin{tabular}{cccllll}
			\toprule
   Category & Occupied & Non-occupied \\
			\midrule
Car & 3.08\% & 96.91\% \\
Chair & 2.51\% & 97.49\% \\
Airplane & 1.42\% & 98.58\% \\ \midrule
Averaged & 2.34\% & 97.66\% \\
   \bottomrule
			\end{tabular}}
   \vspace{-0.5em}
   \caption{{\bf Ratio Statistics} on occupied (foreground) and non-occupied (background) voxels for different categories.
  A significant ratio gap between foreground and background voxels exists.}
 \vspace{-1.5em}
	\label{tab: stat_ratio}
\end{table}

\noindent\textbf{Voxelized Point Clouds Masking.}
For a voxel of resolution $V \times V \times V$ with a total length of $L=(V /p)^3$, we apply a foreground-background masking mechanism to selectively filter out a substantial portion of patches, allowing only the remaining unmasked patches to proceed to the diffusion transformer encoder. Our observations reveal a significant ratio disparity between occupied and non-occupied voxels, as depicted in Table~\ref{tab: stat_ratio}. Considering that occupied voxels contain information richness while background voxels are information-poor, we propose treating voxels in the occupied and non-occupied regions differently to optimize the masking ratio and attain the highest training efficiency. 
Specifically, we apply a ratio of $r_f$ and a ratio of $r_b$ to mask foreground patches $\mathbf{s}_f\in\mathbb{R}^{L_f\times3}$ in occupied voxels and background patches $\mathbf{s}_b\in\mathbb{R}^{L_b\times3}$ in non-occupied voxels, respectively.
Therefore, we only pass $L_u=L-\lfloor r_fL_f\rfloor - \lfloor r_bL_b\rfloor$ unmasked patches to the diffusion transformer encoder.
Our masking approach differs from random masking in image-based diffusion transformers~\cite{zheng2023fast}.
Meanwhile, we empirically observe that the direct extension of MaskDiT~\cite{zheng2023fast} on point clouds does not work well, 
as random masking cannot select meaningful voxels for feature aggregation during the denoising process.
Benefit from the masking strategy, our method is remarkably efficient that an extreme masking ratio $r_b$ (\textit{i.e.}, 99\%) of background patches could still achieve efficient denoising for diffusion steps because the non-occupied background is 97.66\% of overall voxels of all three categories on average, as shown in Table~\ref{tab: stat_ratio}.

\noindent\textbf{Encoder Blocks with 3D Global Attention.}
For encoding point-voxel representations from all unmasked patches $L_u$, we apply multiple encoder blocks based on the global multi-head self-attention operators with each of the heads $\mathbf{Q}, \mathbf{K}, \mathbf{V}$ having dimensions $L_u\times D$, where $L_u$ is the length of input unmasked tokens. 
The global attention operator is formulated as:
    $\mbox{Attention}(\mathbf{Q},\mathbf{K},\mathbf{V}) = \mbox{Softmax}(\dfrac{\mathbf{Q}\mathbf{K}^\top}{\sqrt{D_h}}\mathbf{V})$,
where $D_h$ denotes the dimension size of each head. 
With our extremely high masking ratio,
$L_u$ is 327, while $L$ is 32,768 for $128\times 128\times 128$ input voxels. Thus, given $L_u\ll L$, the computational complexity will be largely reduced to $\mathcal{O}(L_u^2)$ for this encoding process compared to the original complexity $\mathcal{O}(L^2)$ for high voxel resolutions. The efficiency further improves when considering the use of higher-resolution voxel input.

\noindent\textbf{Decoder Blocks with 3D Window Attention.}
During the decoding process, we need to take all encoded unmasked tokens and masked tokens together, which leads to highly expensive complexity $\mathcal{O}(L^2)$ on the increased token length in 3D space. The computational cost of 3D Transformers can be significantly high.
To alleviate this challenge, we are inspired by the original DiT-3D~\cite{mo2023dit3d} and introduce efficient 3D window attention into decoder blocks to propagate point-voxel representations for all input patch tokens using efficient memory.

Specifically, we use a window size $R$ to reduce the length of total input tokens $\hat{P}$ as follows. We first reshape $\hat{P}$ as: $\hat{P} : L\times D \rightarrow \frac{L}{R^3} \times ( D \times R^3 ) $. And then apply a linear layer $\mbox{Linear}(C_{in}, C_{out})(\cdot)$ to $\hat{P}$ : 
    $ P = \mbox{Linear}(D \times R^3, D)(\hat{P}) $.
And $P$ denotes the reduced input patch tokens with a shape of $\frac{L}{R^3}\times D$.
Therefore, the complexity of this decoding process is reduced from $\mathcal{O}(L^2)$ to $\mathcal{O}(\frac{L^2}{R^3})$.

\noindent\textbf{Training Objectives.}
To achieve efficient training using our FastDiT-3D for masked 3D point clouds, we apply a denoising objective $\mathcal{L}_{\text{denoising}}$ on unmasked patches to use a mean-squared loss between the decoder output $\boldsymbol{\epsilon}_{\boldsymbol{\theta}}$ and the ground truth Gaussian noise $\boldsymbol{\epsilon}$, and the objective is simply defined as $\mathcal{L}_{\text{denoising}} = \|\boldsymbol{\epsilon}-\boldsymbol{\epsilon}_{\boldsymbol{\theta}}(\mathbf{x}_{t},t)\|^2$.
To make the model understand the global shape, we also utilize a masked point cloud objective $\mathcal{L}_{\text{mask}}$ on masked patches to minimize the mean-squared loss between the decoder output $\hat{\boldsymbol{\epsilon}}$ and the ground truth Gaussian noise $\boldsymbol{\epsilon}$ at current step $t$ for masked patches. $\mathcal{L}_{\text{mask}} = \|\boldsymbol{\epsilon}-\hat{\boldsymbol{\epsilon}}\|^2$.
Suppose a foreground-background aware mask $\boldsymbol{m}\in\{0,1\}^L$, the overall objective is formulated as,
\begin{equation}
\begin{aligned}
    \mathcal{L} = & E_t (\|(\boldsymbol{\epsilon}-\boldsymbol{\epsilon}_{\boldsymbol{\theta}}(\mathbf{x}_{t},t))\odot(1-\boldsymbol{m})\|^2 + \\
    & \lambda\cdot\|(\boldsymbol{\epsilon}-\hat{\boldsymbol{\epsilon}})\odot\boldsymbol{m}\|^2)
\end{aligned}
\end{equation}
where $E_t ( \| ... \|^2 + \| ... \|^2)$ represents the loss averaged across all timesteps, and $\lambda$ denotes a coefficient to balance the denoising objective and masked prediction. 
In our experiments, we set it to 0.1 in default.
Optimizing the denoising and masked loss together will push the learned representations of our FastDiT-3D to capture global 3D shapes for point cloud generation.

\subsection{Mixture-of-Experts for Multi-class Generation}~\label{sec:moe}
When trained on multi-category point clouds using one single dense model, the generation results will degrade compared to separately trained class-specific models. 
To improve the capacity of multi-category 3D shape generation in a single model, we integrate the Mixture-of-Experts (MoE) design to make the dense model sparse.
Specifically, we replace each encoder block's original Feed Forward Network (FFN) with a MoE FFN.
Given a router network $\mathcal{R}$ and several experts, which formulated as multi-layer perceptions (MLP), $\mathcal{E}_1$, $\mathcal{E}_2$, ..., $\mathcal{E}_n$, where $n$ is the number of experts.
During encoding on the input representations $\mathbf{x}_t$ from different categories, the router $\mathcal{R}$ activates the top-$k$ expert networks with the largest scores $\mathcal{R}(\mathbf{x}_t)_j$, where $j$ denotes the expert index.
In order to sparsely activate different experts, the number of selected experts $k$ is fixed during training and much smaller than the total number of experts $n$.
The expert distribution of our Mixture of Expert (MoE) FFN layers can be formulated as:
\begin{equation}
\begin{aligned}
    \mathcal{R}(\mathbf{x}_t) & = \mbox{TopK}(\mbox{Softmax}(g(\mathbf{x}_t)), k) \\
    \mbox{MoE-FFN}(\mathbf{x}_t) & = \sum_{j=1}^k\mathcal{R}(\mathbf{x}_t)_j\cdot \mathcal{E}_j(\mathbf{x}_t)
\end{aligned}
\end{equation}
where $\mathcal{E}_j(\mathbf{x}_t)$ denotes the representations from the expert $\mathcal{E}_j$, and $g(\cdot)$ is a learnable MLP within the router $\mathcal{R}$.
TopK denotes an operator to select the top $k$ ranked elements with the largest scores from $g(\cdot)$.
By optimizing these experts to balance different categories during training, our FastDiT-3D further achieves adaptive per-sample specialization to generate high-fidelity 3D point clouds for multiple categories.
Each class in this design is capable of capturing a unique diffusion path, involving a variety of experts across various layers. 
This approach significantly eases the challenge of complex gradient optimization that often arises from multi-class joint training.

\begin{table*}[t]
	\renewcommand\tabcolsep{2.0pt}
    \renewcommand{\arraystretch}{1.0}
	\centering
	\scalebox{0.72}{
		\begin{tabular}{l|llll|llll|llll}
		\toprule
\multirow{3}{*}{Method} & \multicolumn{4}{c|}{Chair} & \multicolumn{4}{c|}{Airplane} & \multicolumn{4}{c}{Car} \\
& \multicolumn{2}{c}{1-NNA ($\downarrow$)} & \multicolumn{2}{c|}{COV ($\uparrow$)} & \multicolumn{2}{c}{1-NNA ($\downarrow$)} & \multicolumn{2}{c|}{COV ($\uparrow$)} & \multicolumn{2}{c}{1-NNA ($\downarrow$)} & \multicolumn{2}{c}{COV ($\uparrow$)}\\
& CD  & EMD & CD & EMD & CD & EMD & CD & EMD & CD & EMD & CD & EMD \\
  \midrule
  r-GAN~\cite{achlioptas2018learning}  & 83.69	& 99.70	& 24.27 & 15.13 & 98.40 & 96.79 & 30.12 & 14.32 & 94.46 & 99.01 & 19.03 & 6.539 \\
l-GAN (CD)~\cite{achlioptas2018learning} & 68.58	& 83.84	& 41.99 & 29.31 & 87.30 & 93.95 & 38.52 & 21.23 & 66.49 & 88.78 & 38.92 & 23.58 \\
l-GAN (EMD)~\cite{achlioptas2018learning} & 71.90	& 64.65	& 38.07 & 44.86 & 89.49 & 76.91 & 38.27 & 38.52 & 71.16 & 66.19 & 37.78 & 45.17 \\
PointFlow~\cite{yang2019pointflow} & 62.84	& 60.57	& 42.90 & 50.00 & 75.68 & 70.74 & 47.90 & 46.41 & 58.10 & 56.25 & 46.88 & 50.00 \\
SoftFlow~\cite{Kim2020SoftFlowPF}  & 59.21	& 60.05	& 41.39 & 47.43 & 76.05 & 65.80 & 46.91 & 47.90 & 64.77 & 60.09 & 42.90 & 44.60 \\
SetVAE~\cite{Kim2021SetVAE}  & 58.84	& 60.57	& 46.83 & 44.26 & 76.54 & 67.65 & 43.70 & 48.40 & 59.94 & 59.94 & 49.15 & 46.59 \\
DPF-Net~\cite{Klokov2020dpfnet} & 62.00	& 58.53	& 44.71 & 48.79 & 75.18 & 65.55 & 46.17 & 48.89 & 62.35 & 54.48 & 45.74 & 49.43 \\ 
\midrule
  DPM~\cite{luo2021dpm} & 60.05 & 74.77 & 44.86 & 35.50 & 76.42 & 86.91 & 48.64 & 33.83 & 68.89 & 79.97 & 44.03 & 34.94 \\
  PVD~\cite{zhou2021pvd} & 57.09 & 60.87 & 36.68 & 49.24 & 73.82 & 64.81 & 48.88 & 52.09 & 54.55 & 53.83 & 41.19 & 50.56 \\
LION~\cite{zeng2022lion} & 53.70 & 52.34 & 48.94 & 52.11 & 67.41 & 61.23 & 47.16 & 49.63 & 53.41 & 51.14 & 50.00 & 56.53 \\  
\midrule
GET3D~\cite{gao2022get3d} & 75.26 & 72.49 & 43.36 & 42.77 & -- & -- & -- & -- & 75.26 & 72.49 & 15.04 & 18.38 \\
MeshDiffusion~\cite{liu2023meshdiffusion} & 53.69 & 57.63 & 46.00 & 46.71 & 66.44 & 76.26 & 47.34 & 42.15 & 81.43 & 87.84 & 34.07 & 25.85 \\  
\midrule
DiT-3D-XL~\cite{mo2023dit3d} & \bf 49.11 & 50.73 & 52.45 &  54.32 & 62.35 & 58.67 & 53.16 & 54.39 & 48.24 & 49.35 & 50.00 & 56.38 \\ 
FastDiT-3D-S (ours) & 50.35~\scriptsize{\textcolor{Gray}{(+1.24)}} & \bf 50.27~\scriptsize{\textcolor{ForestGreen}{(-0.46)}}  & \bf 58.53~\scriptsize{\textcolor{ForestGreen}{(+6.08)}} & \bf 60.79~\scriptsize{\textcolor{ForestGreen}{(+6.47)}} &  \bf 61.83~\scriptsize{\textcolor{ForestGreen}{(-0.52)}} & \bf 57.86~\scriptsize{\textcolor{ForestGreen}{(-0.81)}} & \bf 58.21~\scriptsize{\textcolor{ForestGreen}{(+5.05)}} & \bf 58.75~\scriptsize{\textcolor{ForestGreen}{(+4.36)}} & \bf 47.81~\scriptsize{\textcolor{ForestGreen}{(-0.43)}} & \bf 48.83~\scriptsize{\textcolor{ForestGreen}{(-0.52)}} & \bf 53.86~\scriptsize{\textcolor{ForestGreen}{(+3.86)}} & \bf 59.62~\scriptsize{\textcolor{ForestGreen}{(+3.24)}} \\ 
\bottomrule
			\end{tabular}}
\vspace{-0.5em}
\caption{{\bf Comparison results} (\%) on shape metrics of our FastDiT-3D and state-of-the-art models.
Our method significantly outperforms previous baselines in terms of all classes.}
 \label{tab: exp_sota}
\vspace{-1.0em}
\end{table*}

\subsection{Relationship to MaskDiT~\cite{zheng2023fast}}\label{sec:scale}

Our FastDiT-3D contains multiple different and efficient designs for 3D shape generation compared with MaskDiT~\cite{zheng2023fast} on 2D image generation:
\begin{itemize}
    \item We utilize a foreground-background aware masking mechanism with an extremely high masking ratio of nearly 99\%, while MaskDiT~\cite{zheng2023fast} adopted random masking with a relatively low masking ratio of 50\%.
    \item Our FastDiT-3D  performs efficient denoising on voxelized point clouds, while MaskDiT~\cite{zheng2023fast} needs the latent codes from a pre-trained variational autoencoder as the masked denoising target.
    \item We are the first to propose an encoder-decoder diffusion transformer on masked 3D voxelized point clouds for generating high-fidelity point clouds.
\end{itemize}

\section{Experiments}

\subsection{Experimental Setup}

\noindent\textbf{Datasets.}
Following prior works~\cite{zhou2021pvd, zeng2022lion,mo2023dit3d}, we used ShapeNet~\cite{chang2015shapenet} datasets, specifically focusing on the categories of Chair, Airplane, and Car, to serve as our primary datasets for the task of 3D shape generation.
For a fair comparison with previous methods, we sampled 2,048 points from the 5,000 points provided within the ShapeNet dataset~\cite{chang2015shapenet} for training and testing.
For a fair comparison with previous approaches~\cite{zhou2021pvd, zeng2022lion,mo2023dit3d} on 3D shape generation, we follow the same procedures as outlined in PointFlow~\cite{yang2019pointflow} for data preprocessing, which entails global data normalization applied uniformly across the entire dataset.

\noindent\textbf{Evaluation Metrics.}
For comprehensive comparisons, we adopted the same evaluation metrics called Chamfer Distance (CD) and Earth Mover’s Distance (EMD), as in prior methods~\cite{zhou2021pvd, zeng2022lion,mo2023dit3d},  
These metrics are instrumental in computing two key performance indicators: 1-Nearest Neighbor Accuracy (1-NNA) and Coverage (COV), which serve as primary measures of generative quality.
1-NNA computes the leave-one-out accuracy of the 1-Nearest Neighbor (1-NN) classifier to evaluate point cloud generation performance. 
This metric offers robust insights into the quality and diversity of generated point clouds, with a lower 1-NNA score indicating superior performance.
COV quantifies the extent to which generated shapes match reference point clouds, serving as a measure of generation diversity. 
While a higher COV score is desirable, it's important to note that COV primarily reflects diversity and doesn't directly measure the quality of the generated point clouds. 
Therefore, it's possible for low-quality but diverse generated point clouds to achieve high COV scores.

\noindent\textbf{Implementation.}
Our implementation is based on the PyTorch~\cite{paszke2019PyTorch} framework. 
The input voxel size is set to $32\times 32\times 32\times 3$, where $V=32$ represents the spatial resolution.
We perform weight initialization in accordance with established practices, with the final linear layer initialized to zeros and other weights following standard techniques typically employed in Vision Transformers (ViT)~\cite{Dosovitskiy2021vit}.
The models are trained for a total of 10,000 epochs, utilizing the Adam optimizer~\cite{kingma2014adam} with a learning rate of $1e-4$. 
Additionally, we use a batch size of $128$.
In our experiments, we set the diffusion time steps to $T = 1000$.
By default, we apply a small backbone architecture with a patch size of $p=4$.
Notably, global attention is incorporated into all encoder blocks, while 3D window attention is selectively applied to specific decoder blocks (\textit{i.e.}, 1 and 3).
The total number $n$ of experts is 6 in our MoE experiments.

\begin{figure*}[t]
\centering
\includegraphics[width=0.9\linewidth]{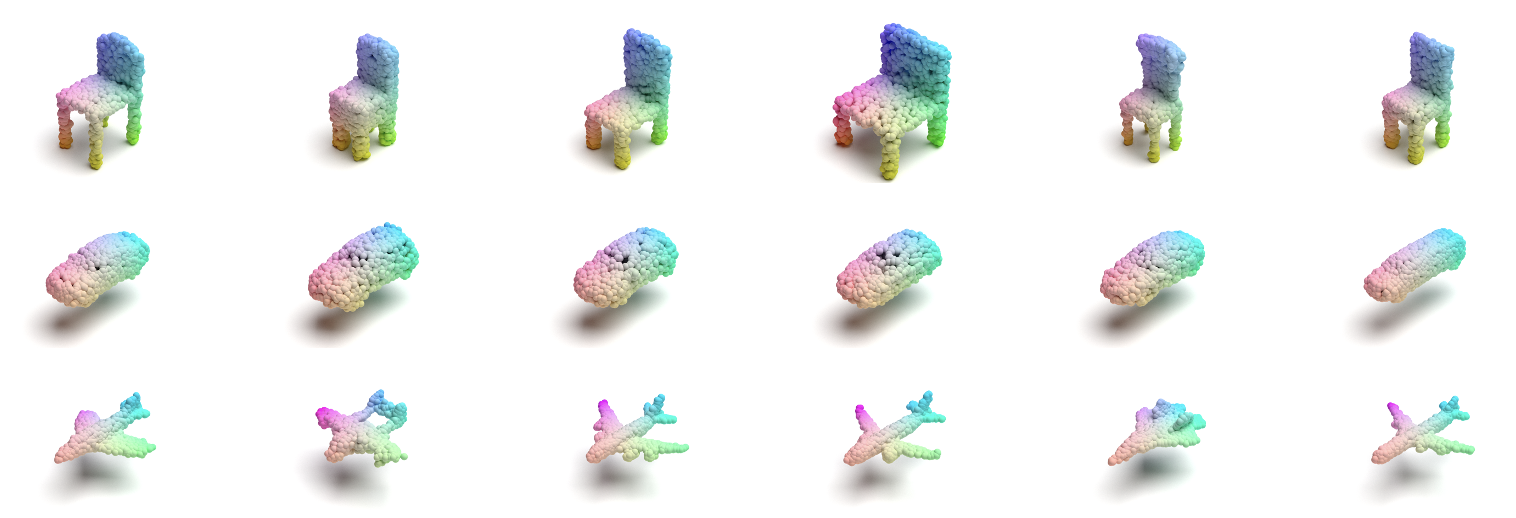}
\vspace{-0.5em}
\caption{{\bf Qualitative visualizations} of high-fidelity and diverse 3D point cloud generation. 
}
\label{fig: vis_generation}
\vspace{-0.5em}
\end{figure*}

\subsection{Comparison to State-of-the-art Works}

In this work, we introduce a novel and highly effective diffusion transformer tailored for 3D shape generation. 
To assess the efficacy of our proposed DiT-3D, we conduct a comprehensive comparative analysis against a range of baseline methods, encompassing both 
non-Diffusion Probabilistic Models (DDPM)~\cite{achlioptas2018learning,yang2019pointflow,Kim2020SoftFlowPF,Kim2021SetVAE,Klokov2020dpfnet,gao2022get3d}, DDPM-based~\cite{luo2021dpm,zhou2021pvd,zeng2022lion,liu2023meshdiffusion}, and Diffusion Transformer-based~\cite{mo2023dit3d} approaches.

We report the quantitative comparison results in Table~\ref{tab: exp_sota}.
As can be seen, we achieved the best results regarding almost all metrics for both 1-NNA and COV evaluations compared to previous 3D shape generation approaches across the three categories.
In particular, the proposed FastDiT-3D in model size of S remarkably superiorly outperforms DiT-3D~\cite{mo2023dit3d} of model size XL, which is the current state-of-the-art diffusion transformer baseline. 

Specifically, our method outperforms DiT-3D for airplane generation, decreasing by 0.52 in 1-NNA@CD and 0.81 in 1-NNA@EMD, and increasing by 5.05 in COV@CD and 4.36 in COV@EMD.
Furthermore, we achieve significant performance gains compared to LION~\cite{zeng2022lion}, a recent competitive baseline based on two hierarchical DDPMs. The results demonstrate the importance of masked prediction in capturing global 3D shapes for point cloud generation.
In addition, significant gains in chair and car generation can be observed in Table~\ref{tab: exp_sota}.
These significant improvements demonstrate the superiority of our approach in 3D point cloud generation.
These qualitative results in Figure~\ref{fig: vis_generation} also showcase the effectiveness of the proposed FastDiT-3D in generating high-fidelity and diverse 3D point clouds.

\begin{table}[t]
	\renewcommand\tabcolsep{6.0pt}
    \renewcommand{\arraystretch}{1.0}
	\centering
	\scalebox{0.8}{
		\begin{tabular}{cccllll}
			\toprule
   3D Voxel & WA  & Training & \multicolumn{2}{c}{1-NNA ($\downarrow$)} & \multicolumn{2}{c}{COV ($\uparrow$)} \\
 Masking & Decoder & Cost (hours) & CD  & EMD & CD & EMD \\
			\midrule
   \xmark  & \xmark &  91 & 51.99 & 50.76 & 54.76 & 57.37 \\
   \cmark  & \xmark &  11 & \bf 50.09 & \bf 50.02 & \bf 59.79 & \bf 61.45 \\
   \cmark  & \cmark & \bf 8 & 50.35 & 50.27 & 58.53 &  60.79 \\
   \bottomrule
			\end{tabular}}
   \vspace{-0.5em}
    \caption{{\bf Ablation studies} on masked 3D components of our FastDiT-3D.
    Our model with both components has the lowest training costs while achieving competitive results.}
 \vspace{-0.5em}
	\label{tab: ab_module}
\end{table}

\begin{table}[t]
	\renewcommand\tabcolsep{6.0pt}
    \renewcommand{\arraystretch}{1.0}
	\centering
	\scalebox{0.85}{
		\begin{tabular}{cccllll}
			\toprule
   \multirow{2}{*}{$r_b$}  & \multirow{2}{*}{$r_f$}  & Training & \multicolumn{2}{c}{1-NNA ($\downarrow$)} & \multicolumn{2}{c}{COV ($\uparrow$)} \\
  & & Cost (hours) & CD  & EMD & CD & EMD \\
			\midrule
\multicolumn{7}{l}{\it Random masking:} \\
\multicolumn{2}{c}{0\%} &  91 & 51.99 & 50.76 & 54.76 & 57.37 \\
\multicolumn{2}{c}{50\%} & 55 & 50.82 & 50.15 & 57.69 & 59.12 \\
\multicolumn{2}{c}{75\%} & 31 & 51.32 & 50.46 & 58.03 & 59.37 \\
\multicolumn{2}{c}{95\%} & 15 & 51.53 & 50.52 & 57.85 & 59.28 \\ 
\multicolumn{2}{c}{99\%} & 11 & 82.35 & 85.16 & 29.63 & 23.56 \\\midrule
\multicolumn{7}{l}{\it Foreground-background aware masking:} \\ 
95\% & 95\% & 15 & 50.22 & 50.06 & 59.25 & 61.23 \\
 97\% & 95\% & 13 & 50.17 & 50.05 & \bf 59.86 & \bf 61.53 \\
99\% & 95\% & 11 & \bf 50.09 & \bf 50.02 & 59.79 &  61.45 \\ \midrule
99\% & \bf 96\% & 10.5 & 50.86 & 50.65 & 57.63 & 58.52 \\
\bf 100\% & 95\% & \bf 10 &  52.87 & 51.69 & 55.23 & 56.82 \\
   \bottomrule
			\end{tabular}}
    \vspace{-0.5em}
    \caption{{\bf Exploration studies} on the trade-off of non-occupied ($r_b$) and occupied ($r_f$) masking ratio.
     When $r_b, r_f$ is 99\%, 95\%, we achieve decent generation results and training costs together.}
	\label{tab: ab_ratio}
   \vspace{-0.5em}
\end{table}

\subsection{Experimental Analysis}

In this section, we performed ablation studies to demonstrate the benefit of introducing two main 3D design components (3D voxel masking and 3D window attention decoder) in 3D shape generation.
We also conducted extensive experiments to explore the efficiency of a mixture-of-experts encoder, modality domain transferability, and scalability.

\noindent\textbf{Ablation on 3D Masked Design Components.}
In order to demonstrate the effectiveness of the introduced 3D voxel masking and 3D window attention (WA) decoder, we ablate the necessity of each module and report the quantitative results in Table~\ref{tab: ab_module}.
We can observe that adding 3D voxel masking to the vanilla baseline highly decreases the training hours from 91 to 11, and improves the generation results by reducing 1.90 in 1-NNA@CD and 0.74 in 1-NNA@EMD and increasing 5.03 in COV@CD and 4.08 in COV@EMD.
Meanwhile, introducing the WA decoder further decreases the training hours to 8, while achieving competitive performance.
These improving results validate the importance of 3D voxel masking and 3D window attention decoder on efficient training and effective 3D point cloud generation.

\newcommand{\modality}{
    \renewcommand{\arraystretch}{1.12}
\begin{tabular}{ccllll}
			\toprule
   ImageNet &  Training  & \multicolumn{2}{c}{1-NNA ($\downarrow$)} & \multicolumn{2}{c}{COV ($\uparrow$)} \\
 Pre-train & Cost (hours) & CD  & EMD & CD & EMD \\
			\midrule
   \xmark  & 8 & \bf 50.35 & \bf 50.27 & 58.53 & 60.79 \\
   \cmark & 7 &  50.39 & 50.28 & \bf 58.62 & \bf 60.86 \\
   \bottomrule
			\end{tabular}
}

\newcommand{\domain}{
\begin{tabular}{ccllll}
			\toprule
   Mixture- & Params  & \multicolumn{2}{c}{1-NNA ($\downarrow$)} & \multicolumn{2}{c}{COV ($\uparrow$)} \\
 of-experts &  (MB) & CD  & EMD & CD & EMD \\
			\midrule
   \xmark &  54.73 & 52.16 & 51.05 & 56.53 & 58.17 \\
   \cmark($k=1$) &  58.26 &   51.95 & 50.87 & 56.86 & 58.63 \\
    \cmark($k=2$) & 68.92 & \bf 51.72 & \bf 50.56 & \bf 57.38 & \bf 59.26 \\
   \bottomrule
			\end{tabular}
}

\begin{table*}[t]
    \centering
    \begin{subfigure}[t]{0.42\textwidth}
        \resizebox{\linewidth}{!}{\modality}
        \caption{Modality transfer.}
       \label{tab: ab_2d}
    \end{subfigure}
    \begin{subfigure}[t]{0.42\textwidth}
        \resizebox{\linewidth}{!}{\domain}
        \caption{Mixture-of-experts. Top $k$ experts are selected.}
        \label{tab: ab_moe}
    \end{subfigure}
    \vspace{-0.5em}
    \caption{{\bf Ablation studies} on 2D pretrain and Mixture-of-experts for multi-category generation. 
    \label{tab: ab_2d_moe}}
\end{table*}

\noindent\textbf{Trade-off of Non-occupied/occupied Masking Ratio.}
The number of non-occupied/occupied masking ratios used in the proposed 3D voxel masking module affects the extracted patch tokens for feature aggregation on point cloud generation.
To explore such effects more comprehensively, we first varied the number of masking ratios from $\{0,50\%,75\%,95\%,99\%\}$ in random masking, and then ablated the non-occupied masking ratio $r_b$ from $\{95\%,97\%,99\%,100\%\}$ and occupied masking ratio $r_f$ from $\{95\%,96\%\}$.
It should be noted that we do not discriminate non-occupied/occupied voxels for random masking, resulting in the same ratio for all voxels.
The comparison results of chair generation are reported in Table~\ref{tab: ab_ratio}.
When the number of masking ratio is 99\% for random masking, we achieve the lowest training costs but the model does not work.
With the increase of non-occupied masking ratio $r_b$ from $95\%$ to $99\%$, the proposed FastDiT-3D consistently improves results in terms of generation quality. The superior performance on such an extreme masking ratio demonstrates the importance of foreground-background aware masking strategy which effectively optimizes the masking ratio and obtains the highest training efficiency. Moreover, we conduct experiments of increasing the non-occupied masking ratio $r_b$ from $99\%$ to $100\%$ and increasing the occupied masking ratio $r_b$ from $95\%$ to $96\%$, the results will not continually improve. 
This is because there might be indispensable voxel patches in both foreground and background for generating high-fidelity point clouds.

\begin{figure}[t]
\centering
\includegraphics[width=0.95\linewidth]{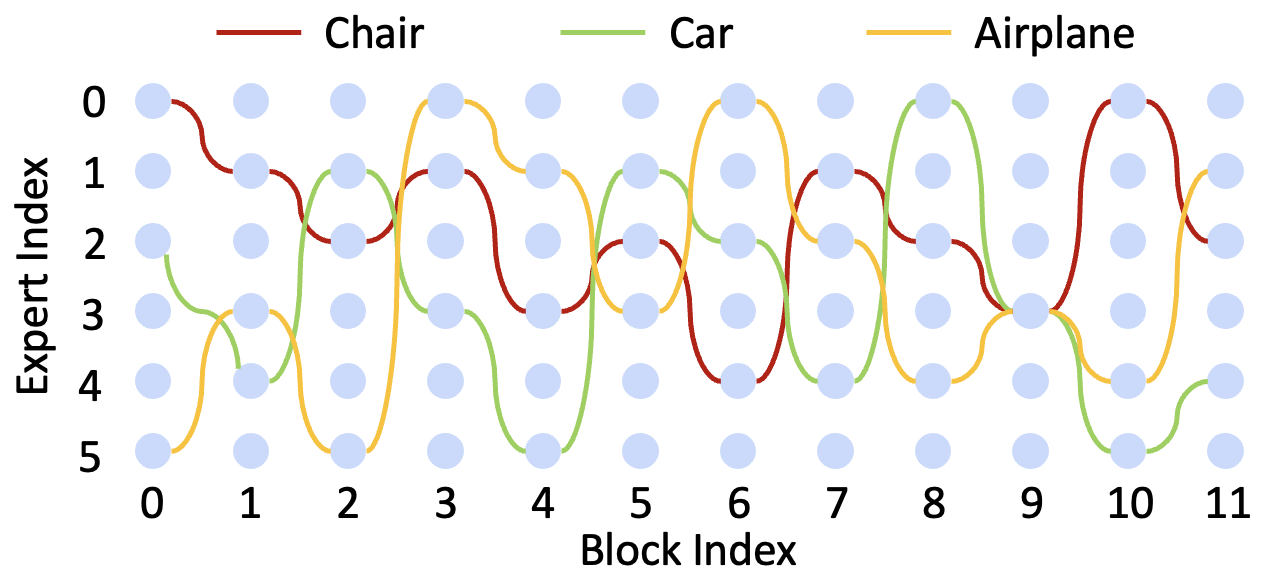}
\vspace{-0.5em}
\caption{{\bf Qualitative visualizations} of sampling paths across experts in Mixture-of-Experts encoder blocks for multi-class generation. 
The learned various paths denote different classes. It demonstrates that each category can learn a distinct unique diffusion path.
}
\label{fig: vis_moe}
\vspace{-1.0em}
\end{figure}

\noindent\textbf{Influence of 2D Pretrain~(ImageNet).}
2D ImageNet pre-trained weights has been demonstrated effective in DiT-3D~\cite{mo2023dit3d} for modality transferability to 3D generation with parameter-efficient fine-tuning.
In order to explore such an effect of modality transferability on our FastDiT-3D, we initialized our encoder and decoder weights from MaskDiT~\cite{zheng2023fast} and continued to fine-tune all parameters during training.
The ablation results on chair generation are reported in Table~\ref{tab: ab_2d}.
We can observe that using ImageNet pre-trained weights achieves fast convergence with fewer training hours and competitive results on high-fidelity point cloud generation, where it outperforms the original random initialization on COV metrics for generating diverse shapes.

\noindent\textbf{Mixture-of-Experts FFN for Multi-class Generation.}
In order to demonstrate the effectiveness of mixture-of-experts FFN in our encoder blocks for generating high-fidelity point clouds from multiple categories, we varied the number of top selected experts $k$ from $\{1,2\}$, and report the comparison results in Table~\ref{tab: ab_moe}.
As can be seen, adding MoE FFN of one expert activated with similar parameters as our FastDiT-3D without MoE achieves better results in terms of all metrics.
Increasing the number of activated experts further improves the performance but brings more training parameters.
These improving results validate the importance of the mixture-of-experts FFN in generating high-fidelity point clouds.
Figure~\ref{fig: vis_moe} also showcases the sample paths across different experts in MoE encoder blocks for multi-category generation for samples from chair, car, and airplane, where the index with the highest frequency of occurrence of experts in each layer are calculated on all training samples corresponding to each class.
We can observe that each class is able to learn a distinct, unique diffusion path, which dynamically chooses different experts in different layers, improving the model’s capacity to generate multiple categories.

\vspace{-0.2em}

\section{Conclusion}

In this work, we propose FastDiT-3D, a novel fast diffusion transformer tailored for efficient 3D point cloud generation. 
Compared to the previous DiT-3D approaches, Our FastDiT-3D dynamically operates the denoising process on masked voxelized point clouds, offering significant improvements in training cost of merely 6.5\% of the original training cost. And FastDiT-3D achieves superior point cloud generation quality across multiple categories.
Specifically, our FastDiT-3D introduces voxel-aware masking to adaptively aggregate background and foreground information from voxelized point clouds, thus achieving an extreme masking ratio of nearly 99\%.
Additionally, we incorporate 3D window attention into decoder Transformer blocks to mitigate the computational burden of self-attention in the context of increased 3D token length.
We introduce Mixture of Expert (MoE) layers into encoder transformer blocks to enhance self-attention for multi-category 3D shape generation.
Extensive experiments on the ShapeNet dataset demonstrate that the proposed FastDiT-3D achieves state-of-the-art generation results in high-fidelity and diverse 3D point clouds.
We also conduct comprehensive ablation studies to validate the effectiveness of voxel-aware masking and 3D window attention decoder.
Qualitative visualizations of distinct sampling paths from various experts across different layers showcase the efficiency of the MoE encoder in multi-category generation.

\newpage

{
    \small
    \bibliographystyle{ieeenat_fullname}
    \bibliography{reference}
}

\newpage

\clearpage

\appendix

\section*{Appendix}

In this appendix, we provide the following material: 
\begin{itemize}
    \item additional experimental analyses on multiple decoder hyper-parameters and various voxel sizes in Section~\ref{sec:analy},
    \item qualitative visualization for comparison with state-of-the-art methods, various voxel sizes, diffusion process, and more generated shapes in Section~\ref{sec:vis},
    \item a demo to show high-fidelity and diverse point clouds generation in Section~\ref{sec:demo},
    \item additional discussions on limitations and broader impact in Section~\ref{sec:limit}.
\end{itemize}

\newcommand{\depth}{
\begin{tabular}{ccllll}
			\toprule
   Decoder  & Training & \multicolumn{2}{c}{1-NNA ($\downarrow$)} & \multicolumn{2}{c}{COV ($\uparrow$)} \\
 Depth & Cost (hours) & CD  & EMD & CD & EMD \\
			\midrule
   4 & 11 & \bf 50.09 & \bf 50.02 & \bf 59.79 & \bf 61.45 \\
    2 & \bf 7 & 51.26 & 50.85 & 55.63 & 57.28 \\
   \bottomrule
			\end{tabular}
}

\newcommand{\width}{
\begin{tabular}{ccllll}
			\toprule
   Decoder & Training & \multicolumn{2}{c}{1-NNA ($\downarrow$)} & \multicolumn{2}{c}{COV ($\uparrow$)} \\
 Width & Cost (hours) & CD  & EMD & CD & EMD \\
			\midrule
   384 & 11 & \bf 50.09 & \bf 50.02 & \bf 59.79 & \bf 61.45 \\
    192 & 8 & 51.78 & 51.51 & 55.21 & 56.17 \\
   \bottomrule
			\end{tabular}
}

\newcommand{\windowsize}{
\renewcommand{\arraystretch}{1.19}
\begin{tabular}{ccllll}
			\toprule
   Window  & Training & \multicolumn{2}{c}{1-NNA ($\downarrow$)} & \multicolumn{2}{c}{COV ($\uparrow$)} \\
 Size & Cost (hours) &  CD  & EMD & CD & EMD \\
			\midrule
   4 & 8 & \bf 50.35 & \bf 50.27 & \bf 58.53 & \bf 60.79 \\
   2 &  \bf 6 & 51.08 & 50.87 & 56.82 & 58.05 \\
   \bottomrule
			\end{tabular}
}

\newcommand{\windowlayer}{
\renewcommand{\arraystretch}{1.19}
\begin{tabular}{ccllll}
			\toprule
   \# WA  & Training & \multicolumn{2}{c}{1-NNA ($\downarrow$)} & \multicolumn{2}{c}{COV ($\uparrow$)} \\
 Layers & Cost (hours) &  CD  & EMD & CD & EMD \\
			\midrule
 2 & 8 & \bf 50.35 & \bf 50.27 & \bf 58.53 & \bf 60.79 \\
3 & \bf 7 & 51.26 & 51.03 & 56.31 & 57.56 \\
   \bottomrule
			\end{tabular}
}

\begin{table*}[h]
    \centering
    \begin{subfigure}[t]{0.42\textwidth}
    \vspace{0pt}
        \resizebox{\linewidth}{!}{\depth}
        \caption{Decoder depth.}
       \label{tab: ab_depth}
    \end{subfigure} 
    \begin{subfigure}[t]{0.42\textwidth}
    \vspace{0pt}
        \resizebox{\linewidth}{!}{\width}
        \caption{Decoder width.}
        \label{tab: ab_width}
    \end{subfigure}  \\
    \begin{subfigure}[t]{0.43\textwidth}
    \vspace{0pt}
        \resizebox{\linewidth}{!}{\windowsize}
        \caption{Window size.}
        \label{tab: ab_window}
    \end{subfigure}
    \begin{subfigure}[t]{0.42\textwidth}
    \vspace{0pt}
        \resizebox{\linewidth}{!}{\windowlayer}
        \caption{Number of Window Attention Layers.}
        \label{tab: ab_window}
    \end{subfigure}
    \vspace{-1em}
    \caption{{\bf Ablation studies} on decoder depth, width, window sizes, and the number of window attention layers. 
    \label{tab: ab_decoder}}
\end{table*}

\section{Additional Experimental Analyses}\label{sec:analy}

In this section, we perform additional ablation studies to explore the effect of multiple hyper-parameters design in decoder and window attention.
We also conduct additional experiments to demonstrate the advantage of the proposed FastDiT-3D against DiT-3D~\cite{mo2023dit3d} on different voxel sizes in terms of training costs and performance.

\subsection{Multiple Hyper-parameters Design in Decoder}

Multiple hyper-parameters including decoder depth/width, window size, and number of window attention layers, in the 3D window attention decoder are also critical for us to reduce expensive training costs and achieve superior performance.
To explore the impact of those key factors, we ablated the decoder depth from $\{4,2\}$, the decoder width from $\{384,192\}$, the window size from $\{4,2\}$, and the number of window attention layers from $\{2,3\}$.
The quantitative results on chair generation are compared in Table~\ref{tab: ab_decoder}.
As shown in the table, when the decoder depth and decoder width are 4 and 384, our FastDiT-3D without window attention layers achieves the best results while having decent training costs. 
Adding window attention with the window size of 4 and the number of layers of 2 further decreases the training hours and achieves competitive performance.

\begin{table}[t]
	\renewcommand\tabcolsep{6.0pt}
    \renewcommand{\arraystretch}{1.0}
	\centering
	\scalebox{0.73}{
		\begin{tabular}{clccccc}
			\toprule
   \multirow{2}{*}{Method}  & Voxel & Training & \multicolumn{2}{c}{1-NNA ($\downarrow$)} & \multicolumn{2}{c}{COV ($\uparrow$)} \\
  & Size & Cost (hours) & CD  & EMD & CD & EMD \\
			\midrule
   \multirow{2}{*}{32} &  DiT-3D~\cite{mo2023dit3d} & 91 & 51.99 & 50.76 & 54.76 & 57.37 \\
     & FastDiT-3D (ours) & \bf 8 & \bf 50.35 & \bf 50.27 & \bf 58.53 & \bf 60.79 \\ \hline
     \multirow{2}{*}{64} & DiT-3D~\cite{mo2023dit3d} & 319 & 51.22 & 50.52 & 55.25 & 57.52 \\
    & FastDiT-3D (ours) & \bf 25 & \bf 50.29 & \bf 50.23 & \bf 58.57 & \bf 60.83 \\ \hline
    \multirow{2}{*}{128} & DiT-3D~\cite{mo2023dit3d} & 1668 & 50.95 & 50.36 & 56.03 & 58.16 \\
     & FastDiT-3D (ours) & \bf 108 & \bf 50.01 & \bf 50.03 & \bf 59.95 & \bf 62.08 \\
   \bottomrule
			\end{tabular}}
    \caption{{\bf Quantitative results} on various voxel sizes ($32$, $64$, $128$). 
    Our model has the lowest training costs while achieving competitive results, compared to DiT-3D~\cite{mo2023dit3d}, the state-of-the-art approach.}
	\label{tab: ab_voxel}
\end{table}

\subsection{Quantitative Results on Various Voxel Sizes}

 To validate the efficiency and effectiveness of the proposed FastDiT-3D on different voxel sizes, we varied the voxel size $V$ from $\{32, 64, 128\}$, and compared our framework with DiT-3D~\cite{mo2023dit3d}, the state-of-the-art approach for point clouds generation.
 The quantitative comparison results are reported in Table~\ref{tab: ab_voxel}.
 We can observe that when the voxel size is 32, our FastDiT-3D achieves better results than DiT-3D~\cite{mo2023dit3d} in terms of all metrics while using only 8.8\% training GPU hours.
 With the increase in the voxel size, we achieve better generation performance and training gains compared to the strong baseline.
 In particular, the proposed FastDiT-3D improves all metrics in terms of generating 128-resolution voxel point clouds and uses only 6.5\% of the training time in DiT-3D~\cite{mo2023dit3d}, reducing the training time from 1668 A100 GPU hours to 108 A100 GPU hours.
These significant results further demonstrate the efficiency of our method in generating high-fidelity and diverse 3D point clouds.

\section{Qualitative Visualizations}\label{sec:vis}

In order to qualitatively demonstrate the effectiveness of the proposed FastDiT-3D in 3D point clouds generation, we compare the generated point clouds with previous approaches.
Meanwhile, we showcase qualitative visualizations of generated point clouds on the chair category using various voxel sizes. Furthermore, we also visualize the diffusion process of different categories generation from the denoising sampling steps.  
Finally, we provide more visualization of 3D point clouds generated by our approach.

\begin{figure}[t]
\centering
\includegraphics[width=0.8\linewidth]{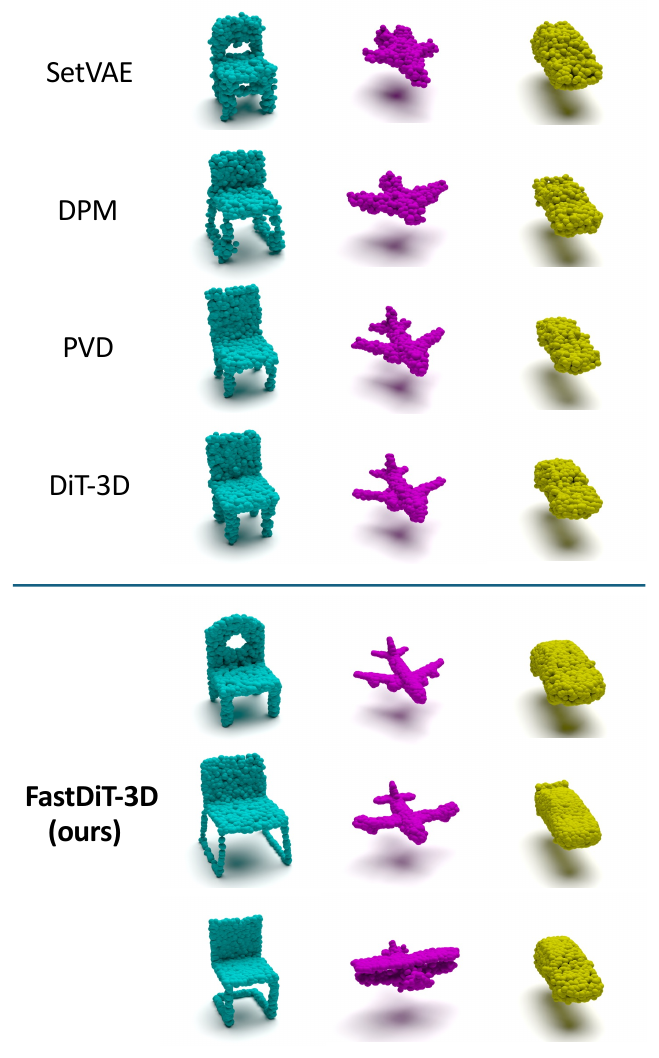}
\vspace{-0.5em}
\caption{{\bf Qualitative comparisons} with state-of-the-art methods for high-fidelity and diverse 3D point cloud generation. 
Our proposed FastDiT-3D produces better results for each category.
}
\label{fig: vis_sota}
\vspace{-0.5em}
\end{figure}

\subsection{Comparisons with State-of-the-art Works}

In this work, we propose a novel framework for generating high-fidelity and diverse 3D point clouds.
To qualitatively demonstrate the effectiveness of the proposed FastDiT-3D, 
we compare our method with previous approaches: 
1) SetVAE~\cite{Kim2021SetVAE}: a hierarchical variational autoencoder for latent variables to learn coarse-to-fine dependencies and permutation invariance; 
2) DPM~\cite{luo2021dpm}): the first denoising diffusion probabilistic models (DDPM) method that applied a Markov chain conditioned on shape latent variables as the reverse diffusion process for point clouds; 
3) PVD~\cite{zhou2021pvd}: a robust DDPM baseline that adopts the point-voxel representation of 3D shapes; 
4) DiT-3D~\cite{mo2023dit3d}: the state-of-the-art diffusion transformer for 3D point cloud generation.

The qualitative visualization results are reported in Figure~\ref{fig: vis_sota}.
As can be observed, 3D point clouds generated by our FastDiT-3D are both high-fidelity and diverse.
The non-DDPM approach, SetVAE~\cite{Kim2021SetVAE}, performs the worst compared to other DDPM methods, although they applied a hierarchical variational autoencoder tailored for coarse-to-fine dependencies.
Furthermore, the proposed framework produces more high-fidelity point clouds compared to DPM~\cite{luo2021dpm} and PVD~\cite{zhou2021pvd} methods.
Finally, we achieve better performance than DiT-3D~\cite{mo2023dit3d} which applied a plain diffusion transformer to aggregate representations from full voxels.
These meaningful visualizations demonstrate the effectiveness of our method in high-fidelity and diverse 3D point clouds generation by adaptively learning background or foreground information from voxelized point clouds with an extreme masking ratio.

\begin{figure*}[t]
\centering
\includegraphics[width=0.8\linewidth]{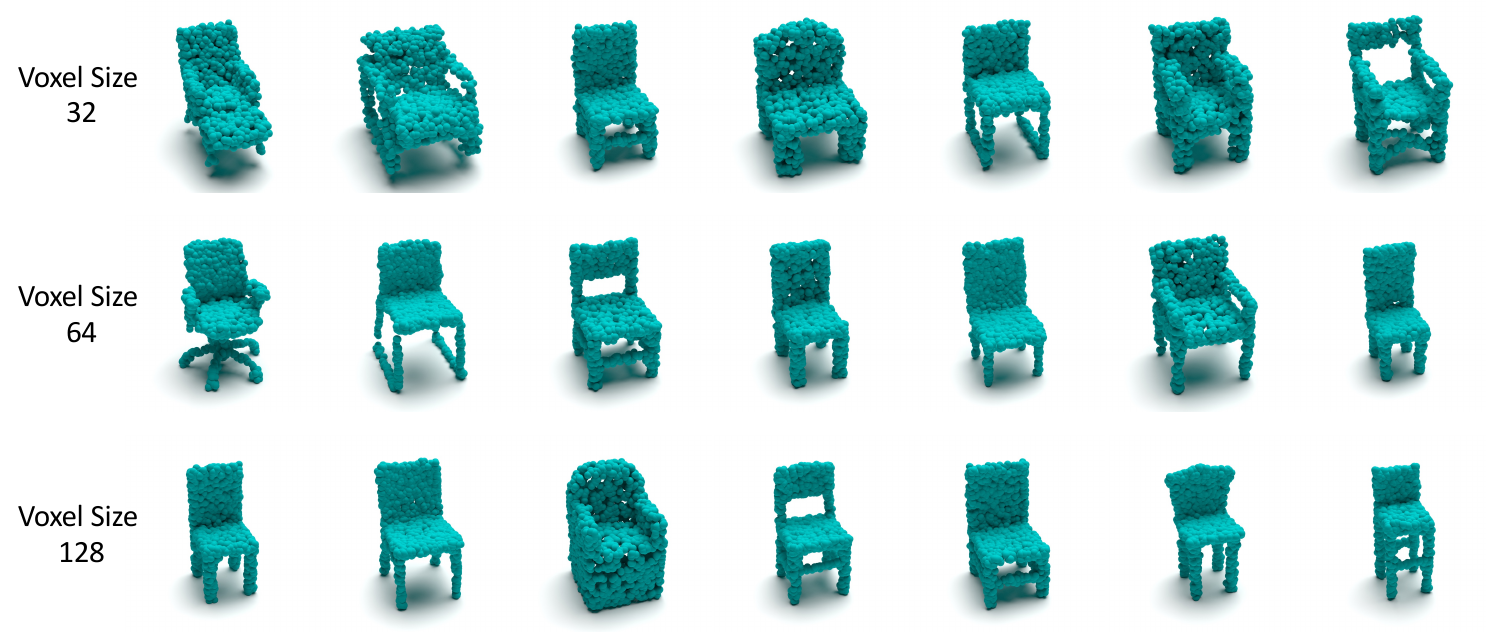}
\caption{{\bf Qualitative visualizations} of generated point clouds on chair category for various voxel sizes.
Rows denote $32$, $64$, and $128$  in top-to-bottom order.
The results showcase the efficiency of our method in generating high-fidelity and diverse 3D point clouds.
}
\label{fig: vis_voxel}
\vspace{-0.5em}
\end{figure*}

\subsection{Various Voxel Sizes}

To validate the effectiveness of our framework in generating high-fidelity and diverse 3D point clouds in different voxel sizes, we visualize generated point clouds in different voxel size from $\{32,64,128\}$ on the chair category in Figure~\ref{fig: vis_voxel}.
As can be seen, with the increase of the voxel size, our FastDiT-3D achieves better results in high-fidelity 3D point clouds generation.
More importantly, the proposed framework produces more fine-grained details when it comes to generating 128-resolution 3D point clouds.
These meaningful qualitative visualizations furthermore show the superiority of our approach in generating high-fidelity 3D point clouds on different voxel sizes.

\subsection{Diffusion Process}

\begin{figure*}[t]
\centering
\includegraphics[width=0.98\linewidth]{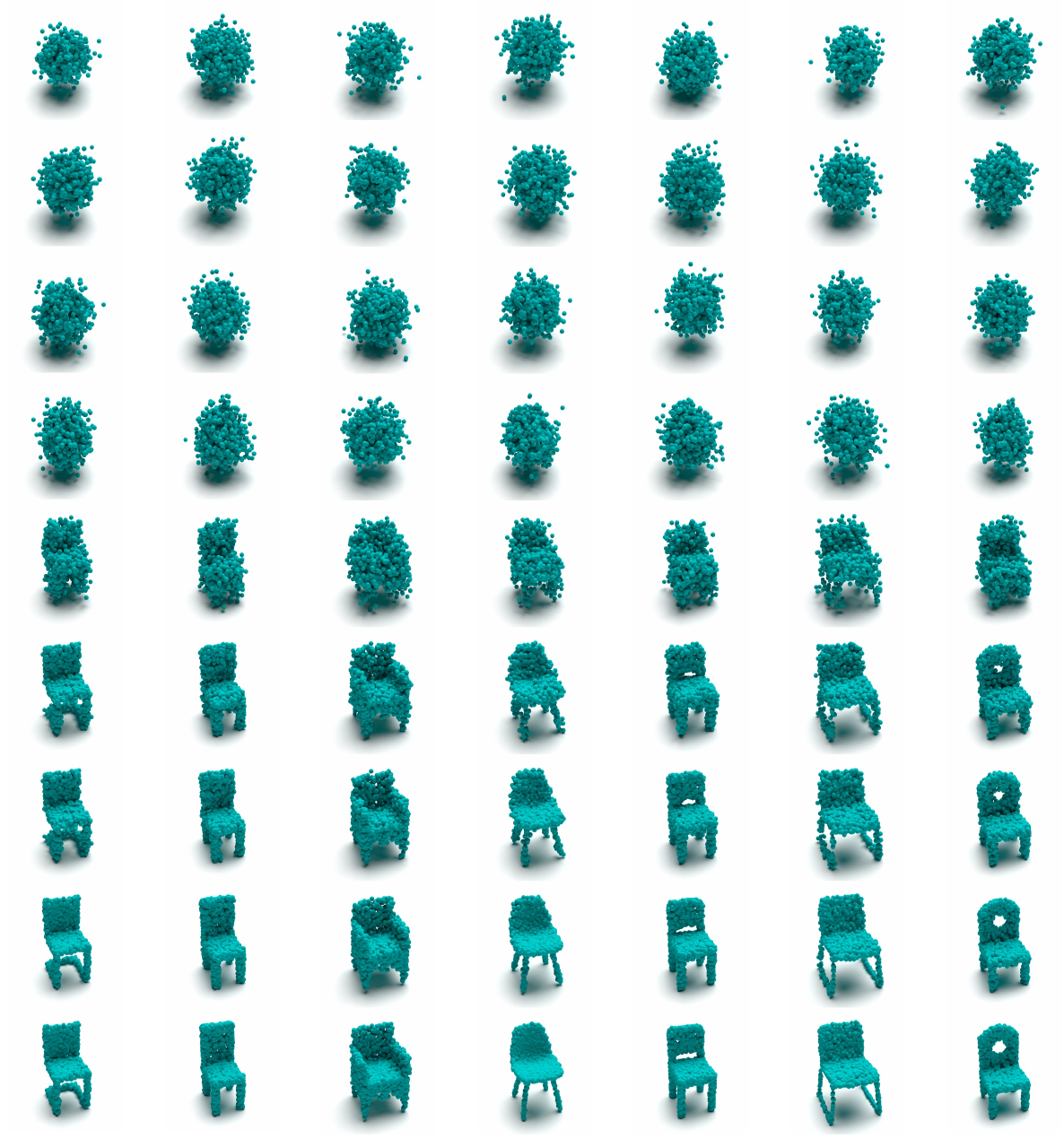}
\caption{{\bf Qualitative visualizations} of diffusion process for chair generation. 
The generation results from random noise to the final 3D shapes are shown in top-to-bottom order in each column.
}
\label{fig: vis_diff_1}
\end{figure*}

\begin{figure*}[t]
\centering
\includegraphics[width=0.98\linewidth]{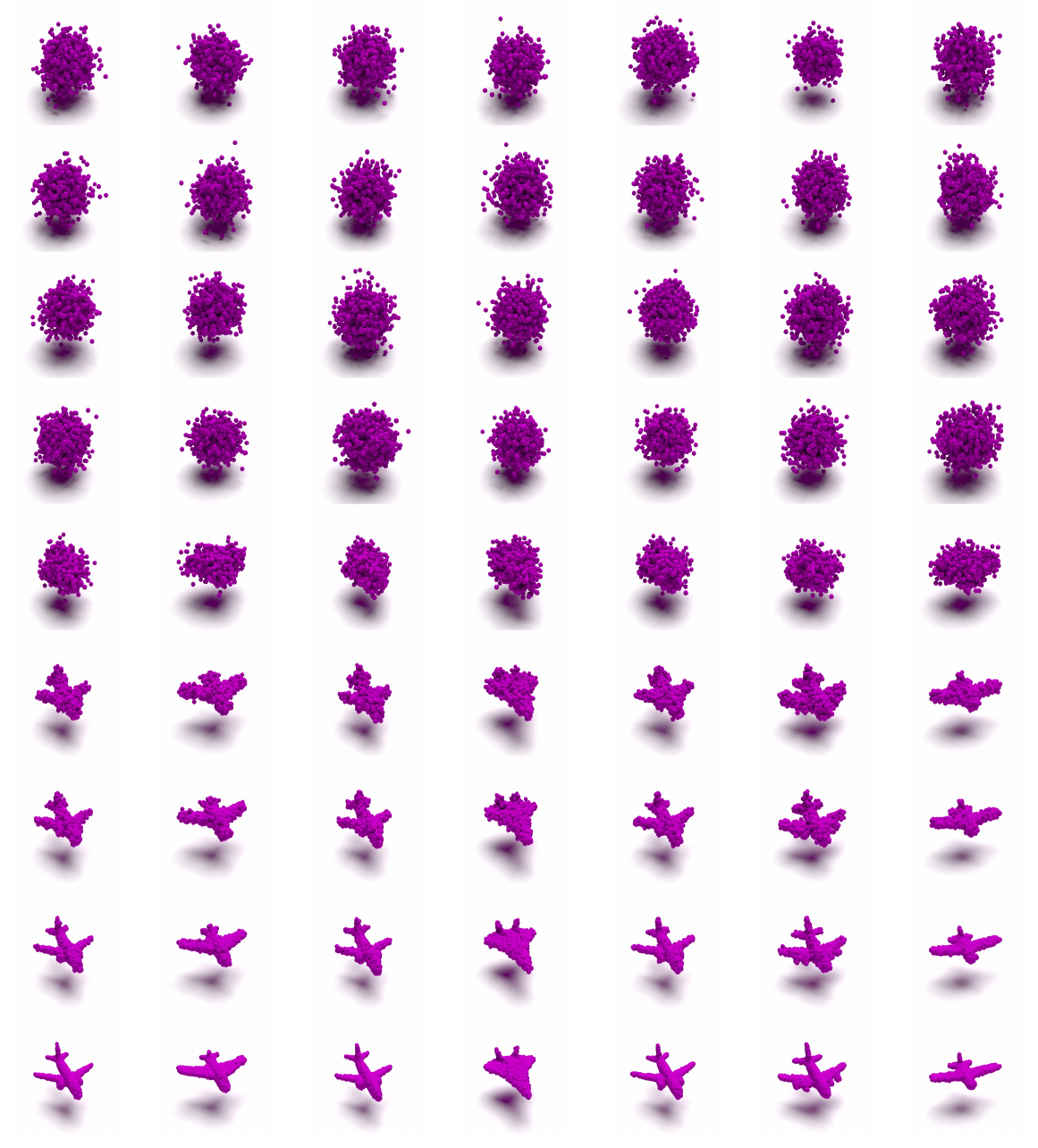}
\caption{{\bf Qualitative visualizations} of diffusion process for airplane generation. 
The generation results from random noise to the final 3D shapes are shown in top-to-bottom order in each column.
}
\label{fig: vis_diff_2}
\end{figure*}

\begin{figure*}[t]
\centering
\includegraphics[width=0.98\linewidth]{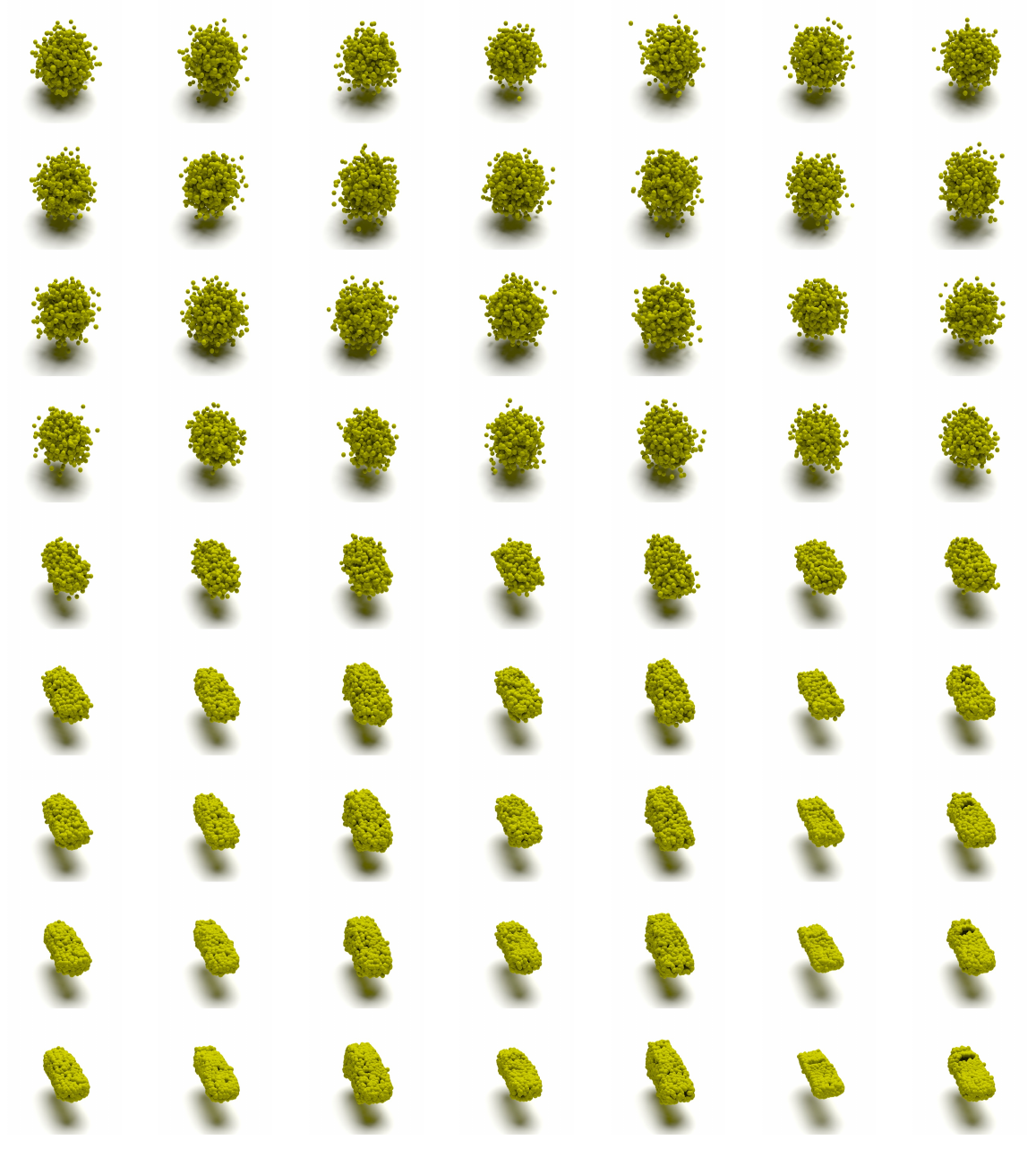}
\caption{{\bf Qualitative visualizations} of diffusion process for car generation. 
The generation results from random noise to the final 3D shapes are shown in top-to-bottom order in each column.
}
\label{fig: vis_diff_3}
\end{figure*}

In order to further demonstrate the effectiveness of the proposed FastDiT-3D, we visualize the diffusion process of generating different categories on 1000 sampling steps.
Specifically, we sample five intermediate shapes in the previous 900 steps and four intermediate shapes in the last 100 steps for better visualization.
Note that for each column, we show the generation results from random noise to the final 3D
shapes in a top-to-bottom order.
Figure~\ref{fig: vis_diff_1} shows the qualitative visualizations of the diffusion process for chair generation, which validates the effectiveness of the proposed FastDiT-3D in generating high-fidelity and diverse 3D point clouds.
The qualitative visualizations of other categories in 
Figure~\ref{fig: vis_diff_2} and~\ref{fig: vis_diff_3} also demonstrate the efficiency of the proposed framework in multi-category generation.

\subsection{More Visualizations of Generated Shapes}

\begin{figure*}[t]
\centering
\includegraphics[width=0.98\linewidth]{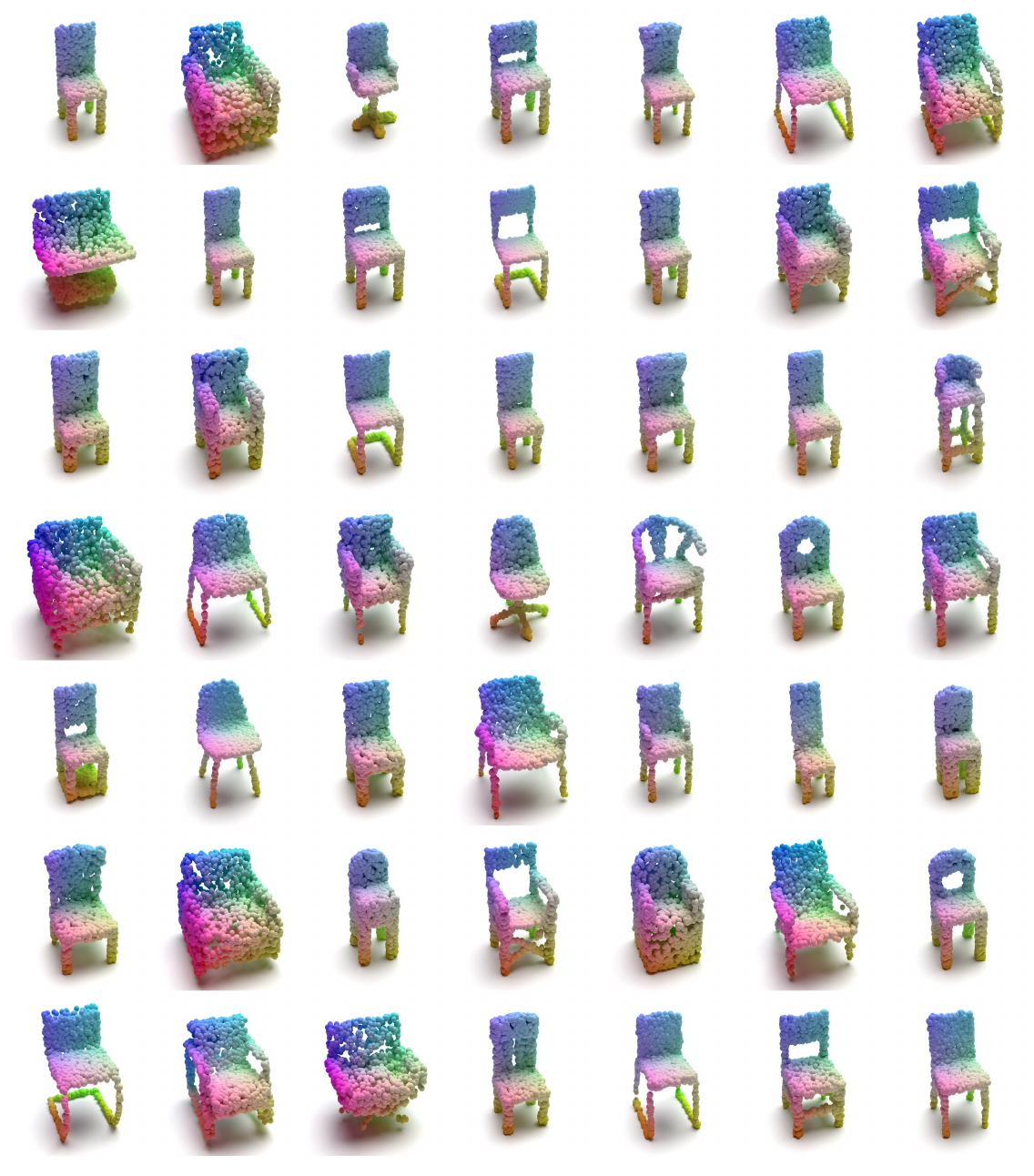}
\caption{{\bf Qualitative visualizations} of more generated shapes on chair category. 
The results showcase the effectiveness of our framework in generating high-fidelity and diverse 3D point clouds.
}
\label{fig: vis_more_1}
\end{figure*}

\begin{figure*}[t]
\centering
\includegraphics[width=0.98\linewidth]{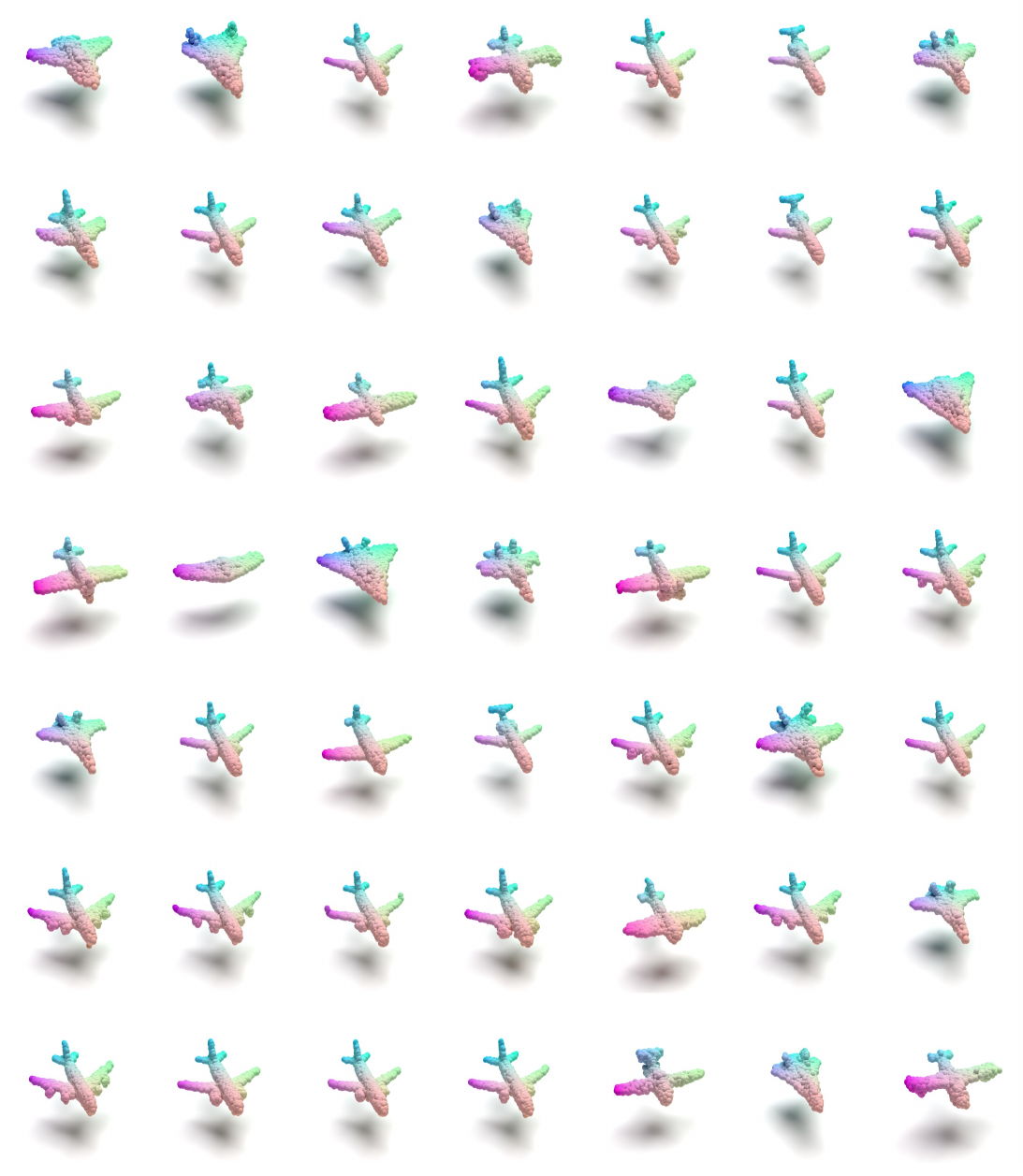}
\caption{{\bf Qualitative visualizations} of more generated shapes on airplane category.
The results showcase the effectiveness of our framework in generating high-fidelity and diverse 3D point clouds.
}
\label{fig: vis_more_2}
\end{figure*}

\begin{figure*}[t]
\centering
\includegraphics[width=0.98\linewidth]{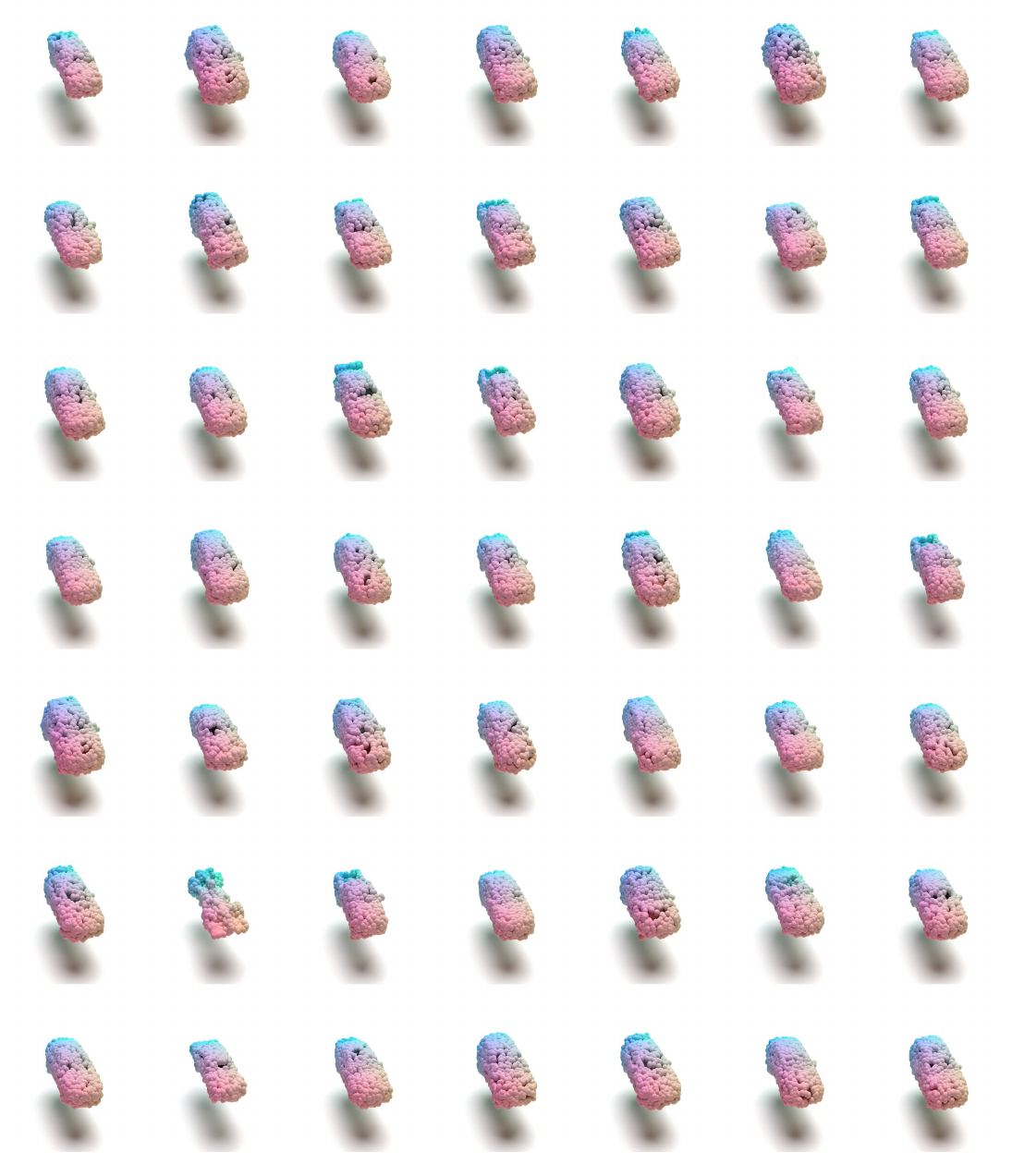}
\caption{{\bf Qualitative visualizations} of more generated shapes on car category. 
The results showcase the effectiveness of our framework in generating high-fidelity and diverse 3D point clouds.
}
\label{fig: vis_more_3}
\end{figure*}

To further validate the effectiveness of our method in generating high-fidelity and diverse 3D point clouds, we visualize more qualitative results generated by our FastDiT-3D from chair, airplane, and car categories in Figure~\ref{fig: vis_more_1},~\ref{fig: vis_more_2}, and~\ref{fig: vis_more_3}.
These meaningful results from different categories further showcase the effectiveness of our framework in generating high-fidelity and diverse 3D point clouds.

\section{Demo}
\label{sec:demo}

The demo is available at our website:~\url{https://DiT-3D.github.io/FastDiT-3D}.

\section{Limitations \& Broader Impact}\label{sec:limit}

Although the proposed FastDiT-3D achieves superior results in generating high-fidelity and diverse point clouds given classes, we have not explored the potential usage of explicit text control for 3D shape generation.
Furthermore, we can scale our FastDiT-3D to large-scale text-3D pairs~\cite{deitke2022objaverse,deitke2023objaversexl} for efficient training on text-to-3D generation.
These promising directions will leave for future work.

\end{document}